\documentclass[conference]{IEEEtran}
\IEEEoverridecommandlockouts
\usepackage{cite}
\usepackage{amsmath,amssymb,amsfonts}
\usepackage{algorithmic}
\usepackage[colorlinks,bookmarksopen,bookmarksnumbered,citecolor=green]{hyperref}

\usepackage{graphicx}
\usepackage{subcaption}
\usepackage{textcomp}
\usepackage{multirow}
\usepackage{booktabs}
\usepackage[table,xcdraw]{xcolor}
\usepackage{float}
\usepackage{utfsym}
\usepackage{array}
\usepackage{threeparttable}

\def\BibTeX{{\rm B\kern-.05em{\sc i\kern-.025em b}\kern-.08em
    T\kern-.1667em\lower.7ex\hbox{E}\kern-.125emX}}
    
\begin{document}

\title{SN-LiDAR: Semantic Neural Fields for Novel Space-time View LiDAR Synthesis\\
\author{
    Yi~Chen$^{1,2}$, Tianchen~Deng$^{1,2}$, Wentao~Zhao$^{1,2}$, 
        Xiaoning Wang$^{3}$, Wenqian Xi$^{4}$,
        Weidong Chen$^{1,2}$, \\
        Jingchuan Wang$^{1,2}$*
\thanks{This work is supported by Shanghai 2024 "Science and Technology Innovation Action Plan" Special Project on Elderly Care Technology Support 24YL1900800.}
\thanks{$^{1}$Department of Automation; Institute of Medical Robotics, Shanghai Jiao Tong University, Shanghai 200030, China}
\thanks{$^{1}$Key Laboratory of System Control and Information Processing of Ministry of Education, Shanghai 200030, China}
\thanks{$^{3}$Ruijin Hospital, Shanghai Jiao Tong University School of Medicine, China} 
\thanks{$^{4}$Renji hospital, Shanghai Jiao Tong University School of Medicine, China}
\thanks{The first two authors contribute equal to this paper.}
\thanks{*Corresponding Author: jchwang@sjtu.edu.cn.}
}
}
% \IEEEauthorblockA{\IEEEauthorrefmark{1}Department of Automation and Institute of Medical Robotics, Shanghai Jiao Tong University, Shanghai 200030, China}
% \affil{\textsuperscript{2}Key Laboratory of System Control and Information Processing of Ministry of Education, Shanghai 200030, China}

% \author{
% 	\IEEEauthorblockN{
% 		Michael Shell\IEEEauthorrefmark{1}, 
% 		Homer Simpson\IEEEauthorrefmark{2}, 
% 		James K irk\IEEEauthorrefmark{3}, 
% 		Montgomery Scott\IEEEauthorrefmark{3} 
% 		and Eldon Tyrell\IEEEauthorrefmark{4}} 
% 	\IEEEauthorblockA{\IEEEauthorrefmark{1}School of Ele ctrical and Computer Engineering\\ Georgia Institute of Technology, Atlanta, Georgia 30 332--0250\\ Email: mshell@ece.gatech.edu}
% 	\IEEEauthorblockA{\IEEEauthorrefmark{2}Twentieth Cen tury Fox, Springfield, USA\\ Email: homer@thesimpsons.com}
% 	\IEEEauthorblockA{\IEEEauthorrefmark{3}Starfleet Aca demy, San Francisco, California 96678-2391\\ Telephone: (800) 555--1212, Fax: (888) 555--1212} 
% 	\IEEEauthorblockA{\IEEEauthorrefmark{4}Tyrell Inc., 123 Replicant Street, Los Angeles, California 90210 --4321}
% } 

\maketitle

\begin{abstract}
Recent research has begun exploring novel view synthesis (NVS) for LiDAR point clouds, aiming to generate realistic LiDAR scans from unseen viewpoints. However, most existing approaches do not reconstruct semantic labels, which are crucial for many downstream applications such as autonomous driving and robotic perception. Unlike images, which benefit from powerful segmentation models, LiDAR point clouds lack such large-scale pre-trained models, making semantic annotation time-consuming and labor-intensive. To address this challenge, we propose SN-LiDAR, a method that jointly performs accurate semantic segmentation, high-quality geometric reconstruction, and realistic LiDAR synthesis. Specifically, we employ a coarse-to-fine planar-grid feature representation to extract global features from multi-frame point clouds and leverage a CNN-based encoder to extract local semantic features from the current frame point cloud. Extensive experiments on SemanticKITTI and KITTI-360 demonstrate the superiority of SN-LiDAR in both semantic and geometric reconstruction, effectively handling dynamic objects and large-scale scenes. Codes will be available on \href{https://github.com/dtc111111/SN-Lidar}{https://github.com/dtc111111/SN-Lidar}.
\end{abstract}

% \begin{IEEEkeywords}
% component, formatting, style, styling, insert
% \end{IEEEkeywords}

\section{Introduction}
% Novel View Synthesis (NVS) generates views from perspectives that sensors have not captured. This technique can produce a broader range of views and data featuring complex behaviors. In autonomous driving systems, it can synthesize rare corner-case scenarios that are seldom recorded. These generated datasets enhance the training and testing of downstream models, leading to improved robustness and generalization. However, the synthesized sensor data requires semantic annotation before it can be applied to perception and planning tasks. This makes the generation of novel views with semantic labels particularly important.

LiDAR Novel View Synthesis (NVS) generates views from perspectives that LiDAR sensors have not captured. This technique can produce a broader range of views and data featuring complex behaviors. In autonomous driving systems, it can synthesize rare corner-case scenarios that are seldom recorded. These generated data improve the training and testing of the downstream models, leading to improved robustness and generalization. 

\begin{figure}[htp]
    \centering
    \includegraphics[width=0.5\textwidth]{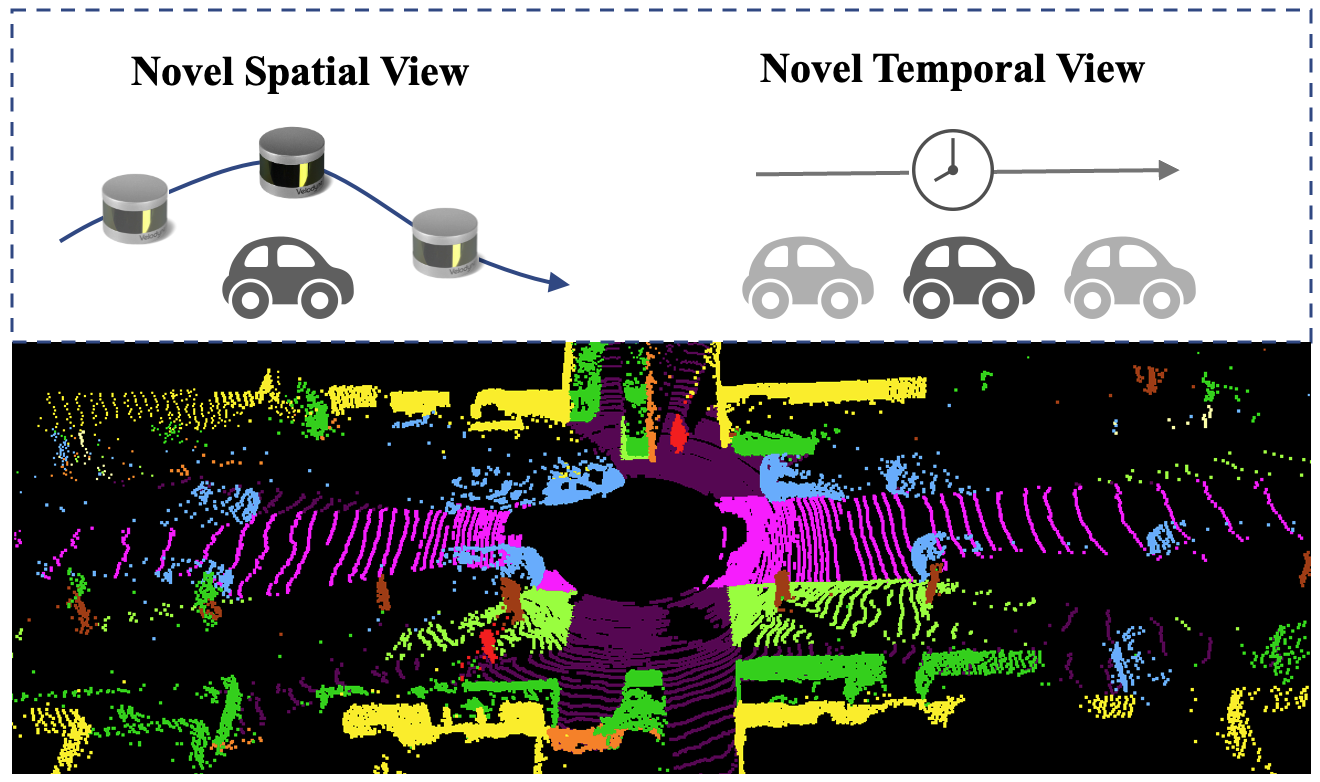}
    \caption{Novel space-time view LiDAR Synthesis with semantics in autonomous driving. Large-scale scenes and dynamic objects are main challenges. }
    \label{fig:main}
\end{figure}

Early solutions for LiDAR NVS are model-based LiDAR simulations \cite{dosovitskiy2017carla} \cite{koenig2004design}. These approaches construct virtual environments and use raycasting to simulate laser sensors, generating LiDAR point clouds from arbitrary viewpoints. However, they require costly 3D assets, and their idealized sensor models lead to a huge domain gap between simulated data and real-world measurements. To address this gap, improved methods \cite{manivasagam2020lidarsim} \cite{li2023pcgen} generate point clouds from real data through a two-step process: first reconstructing 3D scenes from multiple LiDAR scans using surfel \cite{pfister2000surfels} or mesh representations, then casting rays to obtain intersection with surfaces. While model-based LiDAR simulations have made significant strides in generating LiDAR point clouds, these methods rely on explicit reconstruction, which inherently limits their ability to query unscanned points, resulting in challenges for achieving fine-grained geometric reconstruction.  

The introduction of Neural Radiance Fields (NeRF) \cite{mildenhall2021nerf}, with its ability to implicitly reconstruct 3D scenes, has significantly improved synthesis quality and has various applications such as autonomous driving~\cite{prosgnerf}, and robotics localization~\cite{plgslam,neslam} and planning~\cite{planning,navigation}. Therefore, some studies have attempted to adapt NeRF, initially designed for cameras, to LiDAR. Due to fundamental differences between point clouds and images, NeRF cannot be applied directly to LiDAR point clouds. LiDAR data presents challenges: sparse point distribution, discontinuous point patterns, and occlusion between objects. Based on the different principles of how LiDAR sensors measure distance, NFL \cite{huang2023neural} developed a volumetric rendering method suitable for LiDAR considering beam divergence and multiple returns. LiDAR-NeRF \cite{tao2024lidar} takes an image-centric approach by converting LiDAR distance, intensity, and ray-drop attributes into pseudo-images, enabling the use of image-based NeRF methods for LiDAR NVS. To address the limitations of static scene reconstruction, LiDAR4D\cite{zheng2024lidar4d} employs a 4D hybrid feature representation that distinguishes between dynamic and static features, improving its capability to reconstruct dynamic scenes. While these methods represent important first steps in NeRF-based LiDAR NVS, they struggle with scene representation capability when processing dynamic large-scale scenes.   

Furthermore, the synthesized sensor data requires semantic annotation before it can be applied to downstream tasks. Most LiDAR point cloud datasets rely on manually annotated semantic ground truth, which is extremely difficult and expensive. This makes the generation of novel views with semantic labels particularly important.
 
% In the meantime, there has been some works \cite{zhi2021place} \cite{vora2021nesf} \cite{liu2023unsupervised} \cite{zhu2024sni} demonstrating that NeRF can jointly learn geometric and semantic representations. 

To overcome these challenges and achieve semantic reconstruction of urban scenes, we propose a local-to-global feature encoding method, which leverages hierarchical feature extraction to refine local point cloud information progressively and integrates global context through multi-scale representations. Specifically, we combine coarse-to-fine multi-resolution planar-grid features for global representation with local features extracted from the current frame. Second, we propose a fusion of geometric and semantic features to enable mutual enhancement between geometry and semantics. Geometric features provide a better geometric prior for semantic reconstruction, while semantic features offer semantic understanding of dynamic objects for geometric reconstruction. In this way, we can reconstruct geometry and semantics more accurately. 

% To avoid the interference of view-consistent semantic optimization on distance and intensity, we render semantics before incorporating view embeddings into the neural fields.

Overall, we make the following contributions:
\begin{itemize}
    \item We propose SN-LiDAR, the first differential LiDAR-only framework for novel space-time LiDAR view synthesis with semantic labels, which achieves accurate semantic segmentation, high-quality geometric reconstruction, and realistic LiDAR synthesis.
    \item We integrate global geometric features from multi-resolution
planar-grid representation with local semantic features from CNN-based semantic encoder. This fusion method not only strengthens the mutual enhancement between geometry and semantics but also enables processing large-scale scenes from coarse to fine.
    \item Extensive experiments and evaluations on KITTI-360 and SemanticKITTI datasets demonstrate the superiority of our approach in semantic and geometric reconstruction, with the system effectively handling dynamic objects and large-scale scenes.
\end{itemize}

\section{Related Work}
\noindent\textbf{LiDAR Simulation}. Simulating realistic LiDAR data plays a crucial role in training perception models. Model-based simulators, such as CARLA\cite{dosovitskiy2017carla} and \cite{koenig2004design}, use hand-crafted 3D virtual environments and physical models of LiDAR sensors to generate point clouds through ray-casting. These simulators require specific sensor parameters and expensive 3D assets. Although they produce point clouds with precise geometric representations, a significant domain gap exists between simulated and real-world data, which limits their direct use in downstream tasks. To bridge this gap, recent approaches like LiDARsim\cite{manivasagam2020lidarsim} and PCGen\cite{li2023pcgen} reconstruct explicit 3D representations from real-world LiDAR scans. These methods render point clouds using ray-casting and physical LiDAR models, incorporating LiDAR ray-drop patterns to enhance realism. However, explicit reconstruction approaches face challenges in capturing detailed geometry within large-scale complex scenes and cannot generate data for unscanned areas.

\noindent\textbf{NeRF for LiDAR NVS}. With the rapid development of NeRF in image novel view synthesis, some researchers have started exploring NeRF-based methods for LiDAR NVS. Since these methods do not rely on explicit reconstruction, they can synthesize LiDAR point clouds from a wider range of views. LiDAR-NeRF\cite{tao2024lidar} and NFL\cite{huang2023neural} first proposed the task of novel view synthesis for LiDAR sensors. LiDAR-NeRF transforms LiDAR point clouds into range images via cylindrical projection, turning the task into a multi-attribute image NVS problem. It performs neural radiance fields to predict the depth, intensity, and ray-drop probability of points. NFL, on the other hand, explores the physical properties of real laser beams, such as beam divergence and multiple returns, and forms LiDAR volume rendering different from image rendering. Experiments indicate that models for point cloud registration and semantic segmentation, trained with LiDAR point clouds that account for these properties, outperform those trained using model-based simulated point clouds when applied to real-world scenarios. Although these methods produce higher-quality point clouds than model-based approaches, they still introduce artifacts during dynamic object reconstruction. To address this issue, LiDAR4D\cite{zheng2024lidar4d} proposes a 4D hybrid feature representation that separates dynamic and static objects and incorporates scene flow to maintain temporal consistency of the scene geometry. Despite excellent performance, challenges such as long-range vehicle motion and point cloud occlusion remain unresolved.

\noindent\textbf{Semantic NeRF}. Due to the labor-intensive and time-consuming nature of semantic annotation~\cite{salt,sfpnet}, researchers have explored using NeRF to automate the rendering of semantic labels. Semantic-NeRF\cite{zhi2021place} integrates an additional semantic head alongside color and density heads, enabling the estimation of semantics at sampled points. To achieve generic semantic segmentation capability, NeSF\cite{vora2021nesf} trains a multi-scene shared 3D U-Net\cite{cciccek20163d} to encode the pre-trained density field of NeRF while simultaneously training a semantic MLP to decode the features into semantic information. NeSF achieves its generalization ability through training on an extensive dataset with semantic labels, which necessitates high-quality label annotations.  To reduce dependence on precise pixel-level semantic labels, \cite{liu2023unsupervised} designs a self-supervised semantic segmentation framework, which includes a segmentation model continuously trained across different scenes and a corresponding Semantic-NeRF\cite{zhi2021place} model for each scene. The segmentation model provides pseudo ground truth for Semantic-NeRF, and the consistency of Semantic-NeRF is used to refine the semantic labels. SNI-SLAM\cite{zhu2024sni} and SGS-SLAM\cite{sgsslam} integrates multi-level features of color, geometry, and semantics through feature interaction and collaboration, achieving more accurate results, including color rendering, geometric representation, and semantic segmentation.

\begin{figure*}[!t]
    \centering
    \includegraphics[width=\textwidth]{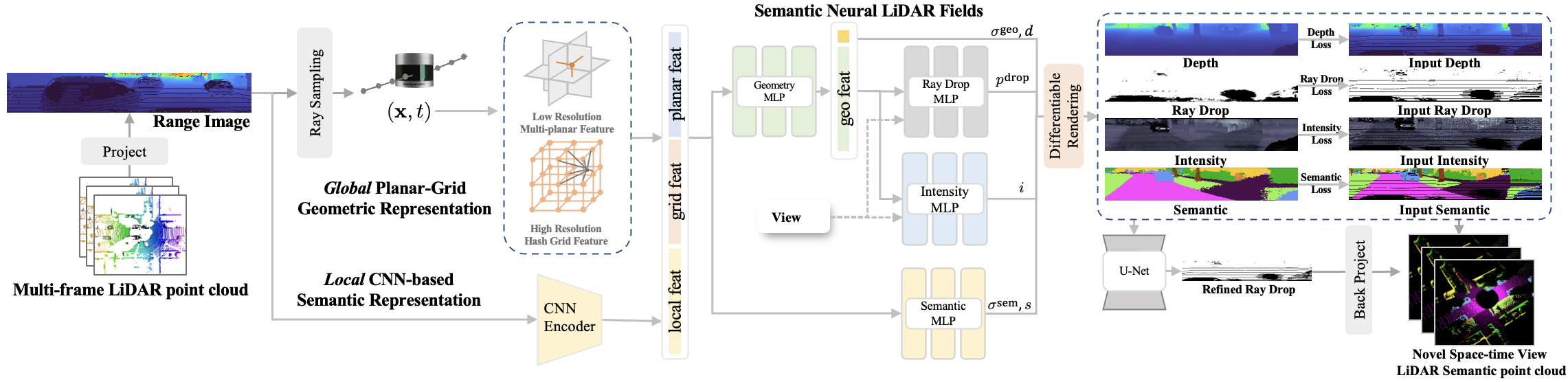}
    \caption{Overall architecture of our proposed SN-LiDAR. For large-scale sparse point clouds in autonomous driving, we combine global geometric and local semantic features within our local-to-global feature representation. The features are fed into semantic neural LiDAR fields for density, intensity, semantic and ray-drop probability prediction. Finally, novel space-time view LiDAR semantic point clouds are synthesized through differentiable rendering and back projecting.}
    \label{fig:architecture}
\end{figure*}

For semantic rendering of LiDAR point clouds, NeRF-LiDAR\cite{zhang2024nerf} uses paired RGB images and LiDAR point clouds as input. The system employs a pre-trained image segmentation model to provide weak label supervision for the images. It then projects these labels from the novel view images onto their corresponding point clouds through cylindrical projection, enabling the generation of semantically labeled point clouds. Therefore, current semantic LiDAR NeRF approaches usually depend on RGB images, with LiDAR data serving only as auxiliary supervision. Our work further explores the semantic rendering of LiDAR-only NeRF, addressing this limitation.

\section{Method}

\subsection{Overall Architecture}\label{sec:overallandarch}
Given a collection of LiDAR scans $\mathcal{X}=\left\{\mathbf{X}_1, \mathbf{X}_2, \ldots,  \mathbf{X}_{NV}\right\} \in \mathbb{R}^{NV \times K \times 5}$, where $\mathbf{X}_n$ contains $K$ points of 3D coordinates $\mathbf{x}=\{x,y,z\}$ , 1D reflection intensity $i$ and 1D semantic label $s$. Scans are associated with sensor poses $P=\left\{P_1, P_2, \ldots, P_{NV}\right\}\left(P_n \in SE(3) \right)$ and timestamps $T=\left\{t_1, t_2, \ldots, t_{NV}\right\}\left(t_n \in \mathbb{R}\right)$. Our goal is to reconstruct the scene as continuous implicit neural fields, from which we could perform neural rendering to synthesize LiDAR point cloud $\mathbf{X}_{novel}$ under any novel sensor pose $P_{novel}$ and time $ t_{novel}$.

The overall architecture of our method is illustrated in Fig.~\ref{fig:architecture}. Given multi-frame LiDAR point clouds as input, we first project them into pseudo range images (Sec.~\ref{sec:rangerepsentation}) to leverage NeRF. To improve scene representation capability when processing dynamic large-scale scenes, we combine global geometric and local semantic features within our local-to-global feature representation module. Sec.~\ref{sec:feaurerepresentation} details the integration of coarse-to-fine multi-resolution plane and grid features to capture global geometry. For local semantic representation, we employ a CNN-based encoder. These planar-grid features and semantic features are then fed into semantic neural LiDAR fields (Sec. \ref{sec:neural}). The features first pass through a geometry MLP to generate fused geometric features and density, which are used for depth rendering. The semantic features are combined with planar-grid features to provide a more comprehensive scene understanding and enable semantic rendering. View embeddings and geometric features are processed by a ray-drop MLP and an intensity MLP to predict ray-drop probabilities and intensities. We also perform global refinement for ray-drop optimization. Finally, rendered pseudo images are back-projected to synthesize novel space-time view LiDAR semantic point clouds. 

\subsection{LiDAR Model and Range Representation}\label{sec:rangerepsentation}
We begin by modeling the LiDAR system, which emits laser beams and measures the time it takes for the beams to hit a reflective surface and return to the sensor. For a LiDAR with $H$ vertical beams and $W$ horizontal emissions, attributes such as depth dd and intensity ii can generate multiple pseudo-images of size $H \times W$. The 3D point coordinates $(x,y,z)$ can be derived from polar coordinates as follows:

\begin{equation}
    \left(\begin{array}{l}x \\y \\z\end{array}\right)=\text{d}\left(\begin{array}{c}\cos (\alpha) \cos (\beta) \\\cos (\alpha) \sin (\beta) \\\sin (\alpha)\end{array}\right)=\text{d} \boldsymbol{\theta}
\end{equation}
where $\alpha$ is the vertical rotation (pitch angle), $\beta$ is the horizontal rotation (yaw angle), $\text{d}(\cdot)$ denotes differential operator, and $\boldsymbol{\theta}$ denotes the viewing direction in the local sensor coordinate system. Specifically, for the 2D coordinates $(h, w)$ in the pseudo range image, we have

\begin{equation}
    \binom{\alpha}{\beta}=\binom{\left|f_{\text {up }}\right|-h f_v H^{-1}}{-(2 w-W) \pi W^{-1}}
\end{equation}
where $f_v=\left|f_{\text {down }}\right|+\left|f_{\text {up }}\right|$ is the vertical field-of-view~(FOV) of the LiDAR sensor, which can be decomposed into downward and upward components $f_{\text{down}}$ and $f_{\text{up}}$. Conversely, each 3D point $(x, y, z)$ in a LiDAR frame is projected onto a pseudo range image of size  $H \times W$ as

\begin{equation}
    \binom{h}{w}=\binom{\left(1-\left(\arcsin (z, d)+\left|f_{\text {down }}\right|\right) f_v^{-1}\right) H}{\frac{1}{2}\left(1-\arctan (y, x) \pi^{-1}\right) W}
\end{equation}
where depth $d$ is calculated as $d=\sqrt{x^2+y^2+z^2}$.

Note that if more than one point projects to the same pseudo-pixel, only the point with the smallest distance is kept. Pixels with no projected points are filled with zeros. In addition, the range image can encode other point attributes, such as intensity.

\subsection{Local-to-Global Feature Representation}\label{sec:feaurerepresentation}
To improve scene representation capability in dynamic large-scale scenarios, we extract features at both global and local scales.

\textbf{Global Planar-Grid Geometric Representation}. Due to the sparse distribution of LiDAR point clouds, directly using dense hash grid features proposed in Instant-NGP\cite{muller2022instant} would lead to redundant memory usage and low efficiency, limiting the scalability to large-scale scenes. Therefore, at the global scale, we employ a coarse-to-fine feature representation to efficiently store the sparse point cloud features. Specifically, we follow LiDAR4D \cite{zheng2024lidar4d}, which combines low-resolution multi-plane features with high-resolution hash grid features. 

The multi-plane features follow K-Planes\cite{fridovich2023k}, which decompose the scene space into a combination of multiple orthogonal planes, significantly reducing number of parameters. The plane features are obtained as follows:
\begin{equation}
    \mathbf{f}_{\text{planar}} = \mathcal{S}(\textbf{V}, (x, y, z, t)), \quad \textbf{V} \in \mathbb{R}^{(3M^2 + 3MH)C}
\end{equation}
where $\textbf{V}$ stores features with $M$ spatial resolution, $H$ temporal resolution and $C$ channels. $\mathcal{S}$ refers to the sampling function that projects 4D coordinates into the corresponding planes $(xy,\ xz,\ yz,\ xt,\ yt,\ zt)$  and interpolates features bilinearly. $(xy, xz, yz)$ are static features while $(xt,\ yt,\ zt)$ stands for dynamic components.

The multi-level grid features follow Instant-NGP\cite{muller2022instant}, which is a high-resolution hash grid structure that enables the handling of fine details of the scene. The grid features are obtained as follows:
\begin{equation}
    \mathbf{f}_{\text{grid}} = \mathcal{S}(\textbf{G}, (x, y, z, t)), \quad \textbf{G} \in \mathbb{R}^{(M^3 + 3M^2H)C}
\end{equation}
where the dense grid $\textbf{G}$ will be further compressed into limited storage via hash mapping for parameter reduction. Similarly, the 4D coordinates are projected into static $(xyz)$ and dynamic $(xyt,\ xzt,\ yzt)$ multi-level hash grids.

\textbf{Local CNN-based Semantic Representation}. To enhance the network's fine-grained perception capability, we extract local semantic features from the current frame point cloud. 

\begin{figure}[htbp]
    \centerline{\includegraphics[width=0.5\textwidth]{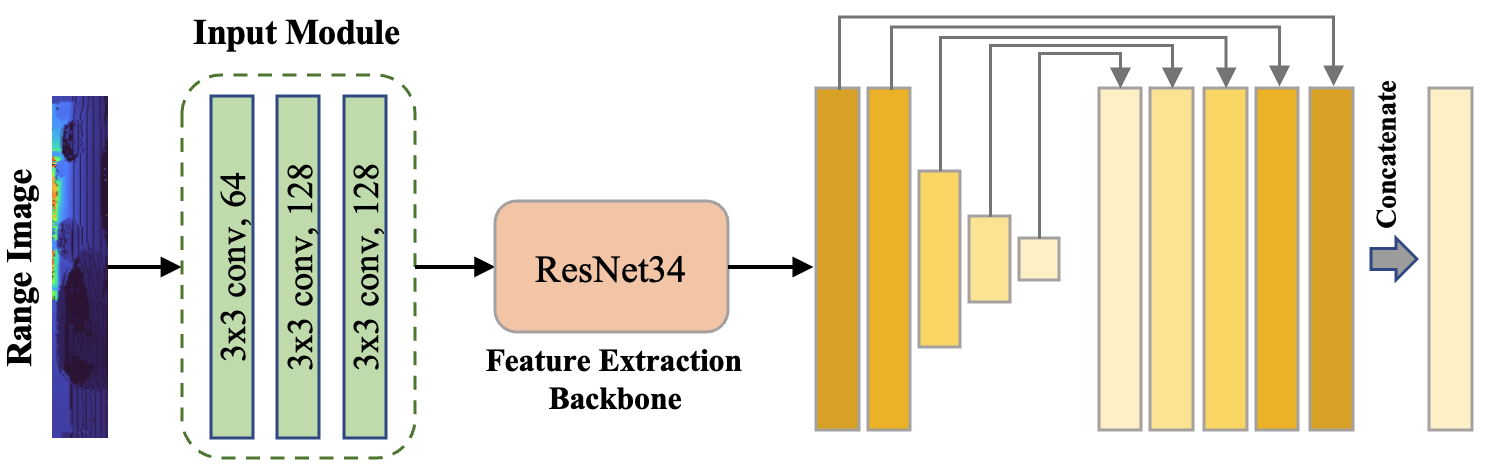}}
    \caption{Local CNN-based semantic encoder. It extracts semantic features for 1-channel range images.}
    \label{fig:encoder} 
\end{figure}

In terms of the encoder network structure, considering that we use the range image as an intermediate representation of point cloud, we follow RangeNet++\cite{milioto2019rangenet++}, which combines convolutional neural networks (CNN) with point cloud semantic segmentation for fast and accurate segmentation. Specifically, we refer to the lightweight range image segmentation network CENet\cite{cheng2022cenet}, which strikes a good balance between network parameters and segmentation performance. We utilize its feature extraction module pretrained on SemanticKITTI and modify the input and output to fit our pipeline, as shown in Fig. \ref{fig:encoder}\ref{fig:encoder}. The network includes an input module composed of $3\times3$ conv layers and a feature extraction backbone, where we choose ResNet34\cite{he2016deep} with Hardswish\cite{howard2019searching} activation functions. Finally, the local semantic features are obtained as follows:
\begin{equation}
     \mathbf{f}_{\text{local}}=\textbf{E}(\mathbf{X}_n)
\end{equation}
where $\mathbf{X}_n$ is the pseudo range image of the n-th frame point cloud, and $\textbf{E}$ represents the CNN Encoder and interpolating features to the image size. The introduction of such features not only enhances the network's ability to capture fine-grained scene details but also enables $\mathbf{f}_{\text{local}}$ to be jointly optimized with $\mathbf{f}_{\text{planar}}$ and $\mathbf{f}_{\text{grid}}$ through the semantic neural fields, facilitating the mutual enhancement of geometry and semantics.

This local-to-global representation efficiently handles point cloud sparsity, reducing memory consumption while maintaining high-quality reconstruction for dynamic and large-scale scenes. By combining local feature enhancement with global contextual awareness, our method enhances both the scalability and accuracy in large environments.

\subsection{Semantic Neural LiDAR Fields}\label{sec:neural}

% Semantic LiDAR NeRF faces two key challenges: (1) The interdependence of distance, intensity, and semantics means that treating them separately fails to provide a complete understanding of the scene; (2) While distance and intensity are view-dependent, semantics remain consistent across perspectives, making joint optimization  could negatively impact their individual performance. 

We propose a differential Semantic Neural LiDAR Fields to jointly decode depth, semantics, intensity and ray drop. The Geometry MLP integrates information from global planar-grid geometric features and local semantic features as input and outputs geometric features and density. The geometric features and view embeddings are subsequently processed by the Ray Drop MLP and Intensity MLP to obtain ray-drop probability and intensity, respectively. Since distance and intensity vary with the viewpoint while semantics remain consistent across different perspectives, joint optimization may negatively impact their individual performance. To preserve semantic consistency across views, the Semantic MLP does not take view embeddings as input and instead leverages planar-grid features and local semantic features. This ensures that semantic optimization does not degrade the synthesis quality of depth and intensity.

During the rendering stage, for each ray  $\mathbf{r}$ emitted from the sensor center $\mathbf{o}$ in direction $\mathbf{d}$, we sample $N$ points $\{p_n\}_{n=1}^N$. The features of 3D sample points are then queried and fed into the neural fields to obtain their attributes and volume densities. The attributes include depth $d_n$, semantics $s_n$, intensity $i_n$, and ray-drop probability $p_n^{\text{drop}}$, while the volume densities consist of geometric density $ \sigma_n^{\text{geo}}$ and semantic density $\sigma_n^{\text{sem}}$.

\begin{equation}
    \text{MLP}_{\text{geometry}} \left( \mathbf{f}_{\text{planar}}, \ \mathbf{f}_{\text{grid}}, \ \mathbf{f}_{\text{local}} \right) \Rightarrow \sigma_n^{\text{geo}},\mathbf{f}_{\text{geo}}
\end{equation}
\begin{equation}
    \text{MLP}_{\text{semantic}} ( \mathbf{f}_{\text{planar}}, \mathbf{f}_{\text{grid}}, \mathbf{f}_{\text{local}}) \Rightarrow \sigma_n^{\text{sem}}, s_n
\end{equation}
\begin{equation}
    \text{MLP}_{\text{intensity}} \left( \mathbf{f}_{\text{geo}}, \ \gamma(\mathbf{d})\right) \Rightarrow i_n
\end{equation}
\begin{equation}
    \text{MLP}_{\text{ray-drop}} \left( \mathbf{f}_{\text{geo}}, \ \gamma(\mathbf{d})\right) \Rightarrow p_n^{\text{drop}}
\end{equation}
where $\gamma(\mathbf{d})$ represents view embeddings:
\begin{equation}
    \gamma(x) = (\sin(2^0 x), \cos(2^0 x), \dots, \sin(2^{L-1} x), \cos(2^{L-1} x))
\end{equation}

Then, depth  $\hat{d}$  can be obtained by integrating density along the ray $\mathbf{r}$:
\begin{equation}
    \hat{d}(\mathbf{r})=\sum_{n=1}^N w_d \cdot d_n
\end{equation}
with
\begin{equation}
    w_d=\text{exp}\left(-\sum_{i=1}^{n-1}  \sigma_i^{\text{geo}} \cdot \delta_i \right) \left(1-e^{ -\sigma_n^{\text{geo}} }\right)
\end{equation}
where $d_n$ is the depth value of queried points on the ray  $\mathbf{r}$, $\delta_i$~is the distance between adjacent samples, and $\alpha_n = 1-e^{ -\sigma_n^{\text{geo}} }$ is opacity. Sharing weights with depth, ray-drop probability $ \hat{p}^{\text{drop}}$ and intensity $\hat{i}$  can be obtained as follows:
\begin{equation}
      \hat{p}^{\text{drop}}(\mathbf{r}) = \sum_{n=1}^N w_d \cdot p_n^{\text{drop}}
\end{equation}
\begin{equation}
    \hat{i}(\mathbf{r})=\sum_{n=1}^N w_d \cdot i_n
\end{equation}
The semantic prediction values $\hat{s}$ can also be obtained through volume rendering of semantic density:
\begin{equation}
    \hat{s}(\mathbf{r})=\sum_{n=1}^N w_s \cdot s_n
\end{equation}
with
\begin{equation}
    w_s=\text{exp}\left(-\sum_{i=1}^{n-1}  \sigma_i^{\text{sem}} \cdot \delta_i \right) \left( 1-e^{ -\sigma_n^{\text{sem}} } \right)
\end{equation}

\subsection{Optimization}\label{sec:optimization}
For the optimization of SN-LiDAR, the total reconstruction loss is the weighted sum of the depth loss, semantic loss, intensity loss and ray-drop loss.

\begin{equation}
    \mathcal{L}_\text{total} = \lambda_{\alpha} \mathcal{L}_{\text{depth}} + \lambda_{\beta} \mathcal{L}_{\text{semantic}} +  \lambda_{\gamma} \mathcal{L}_{\text{intensity}} + \lambda_{\eta} \mathcal{L}_{\text{raydrop}}
\end{equation}
with
\begin{equation}
    \mathcal{L}_{\text{depth}} = \sum_{\mathbf{r} \in R} \left\| \hat{d}(\mathbf{r}) - d(\mathbf{r}) \right\|_1
\end{equation}
\begin{equation}
    \mathcal{L}_{\text{semantic}} = \sum_{\mathbf{r} \in R} s(\mathbf{r}) \cdot \text{log} \hat{s}(\mathbf{r})
\end{equation}
\begin{equation}
    \mathcal{L}_{\text{intensity}} = \sum_{\mathbf{r} \in R} \left\| \hat{i}(\mathbf{r}) - i(\mathbf{r}) \right\|_2^2
\end{equation}
\begin{equation}
    \mathcal{L}_{\text{raydrop}} = \sum_{\mathbf{r} \in R} \left\| \hat{p}^{\text{drop}}(\mathbf{r}) - p^{\text{drop}}(\mathbf{r}) \right\|_2^2
\end{equation}
where $R$ is the set of training rays and $\lambda$ are weight coefficients for each term.

\section{Experiments}

\subsection{Experimental Setup}
\textbf{Datasets}. We conducted comprehensive experiments on the public autonomous driving datasets SemanticKITTI\cite{behley2019iccv} and KITTI-360\cite{Liao2022PAMI}.  SemanticKITTI is captured by a 64-beam LiDAR sensor with 360$^\circ$ horizontal FOV and 26.8$^\circ$ vertical FOV at 10Hz. KITTI-360 has a 64-beam LiDAR, 26.4$^\circ$ vertical FOV, and 10Hz acquisition rate. Both of them have ground truth semantic labels. We selected 50 consecutive frames as a single scene, each covering 100m to 200m, and held out every 10-th frame as a test view.

\textbf{Metrics}. To evaluate the quality of the novel LiDAR point cloud, we convert the rendered range image to a point cloud, and then calculate the Chamfer Distance (CD [m]) \cite{fan2017point} and F-Score with a threshold of 5cm CD error.  Chamfer Distance between point clouds $S_1, S_2 \subseteq \mathbb{R}^3$ is computed as

\begin{equation}
    \text{CD}(S_1, S_2) = \sum_{x \in S_1} \min_{y \in S_2} \| x - y \|_2^2 + \sum_{y \in S_2} \min_{x \in S_1} \| x - y \|_2^2
\end{equation}

For depth and intensity reconstruction results, we calculate pixel-by-pixel error of rendered range images with Root Mean Square Error~(RMSE) and Median Absolute Error~(MedAE). Moreover, we measure reconstruction quality using PSNR for pixel-level accuracy, SSIM \cite{wang2004image} for structural similarity, and LPIPS \cite{zhang2018unreasonable} for perceptual quality. For ray-drop probabilities, we calculate pixel-wise error RMSE, Accuracy and F1-Score. Semantic reconstruction is evaluated with respect to Mean Intersection over Union (mIoU) \cite{long2015fully} and Pixel Accuracy (PA) metric. 

% \textbf{Baselines}. We compare the point cloud synthesis quality with model-based LiDAR simulator LiDARsim~\cite{manivasagam2020lidarsim} and PCGen~\cite{li2023pcgen}, NeRF-based methods LiDAR-NeRF~\cite{tao2024lidar} and LiDAR4D~\cite{zheng2024lidar4d}. Since these methods do not reconstruct semantics, we attached a pre-trained point cloud semantic segmentation network behind them to obtain semantic metrics.

\textbf{Implementation Details}. We use 16-channel features to represent geometry and 128-channel for semantics. The decoder MLPs in neural fields have 3 layers, and the hidden layer dimension is 64. We sample 768 points for each ray. All experiments were conducted on a single NVIDIA A40 GPU.

\subsection{Evaluation of LiDAR Novel View Synthesis}
\begin{figure}[htp]
    \centering
    \includegraphics[width=0.49\textwidth]{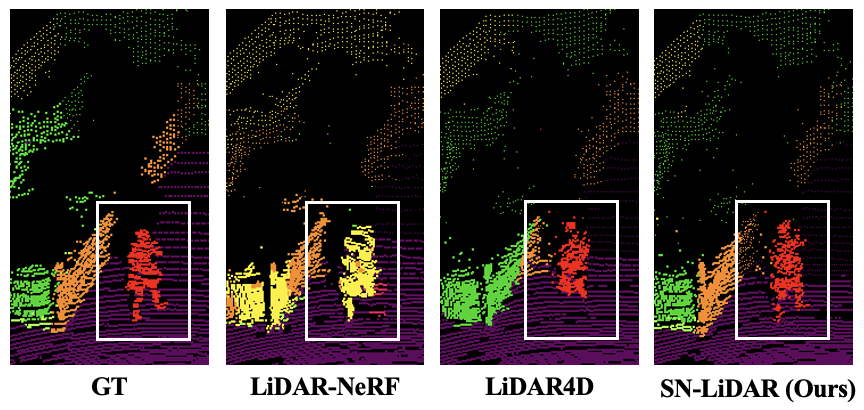}
    \caption{Qualitative comparison for LiDAR \textbf{point cloud} reconstruction and synthesis on SemanticKITTI. The white box shows the point cloud of the pedestrian. }
    \label{fig:pointcloud_semkitti_narrow}
\end{figure}

%%%%%%%%%%%%%%%%%%%%%%%%%%%%%%%%%%%%%%% Tables %%%%%%%%%%%%%%%%%%%%%%%%%%%%%%%%%%%%%%%%%%%
{\renewcommand{\arraystretch}{1.2}
\begin{table*}[!htp]
\centering
\caption{Quantitative comparison on KITTI-360 dataset.}
\label{tab:quantity_KITTI360}
\resizebox{\textwidth}{!}{
\begin{threeparttable}
    \begin{tabular}{cccccccccccccccc}
    \toprule
    \multirow{2}{*}{Method}  & \multicolumn{2}{c}{Point Cloud} & \multicolumn{5}{c}{Depth} & \multicolumn{5}{c}{Intensity} & \multicolumn{2}{c}{Semantic} \\ 
    \cmidrule(lr){2-3} \cmidrule(lr){4-8} \cmidrule(lr){9-13} \cmidrule(lr){14-15} 
                           & CD$\downarrow$ & F-score$\uparrow$ & RMSE$\downarrow$ & MedAE$\downarrow$ & LPIPS$\downarrow$ & SSIM$\uparrow$ & PSNR$\uparrow$ & RMSE$\downarrow$ & MedAE$\downarrow$ & LPIPS$\downarrow$ & SSIM$\uparrow$ & PSNR$\uparrow$  & PA$\uparrow$ & mIoU$\uparrow$ \\ 
    \midrule
    LiDARsim \cite{manivasagam2020lidarsim}  & 3.2228  & 0.7157  & 6.9153  & 0.1279  & 0.2926  & 0.6342  & 21.4608  & 0.1666  & 0.0569  & 0.3276  & 0.3502 & 15.5853 & \textemdash & \textemdash  \\ 
    NKSR \cite{huang2023nksr}             & 1.8982  & 0.6855  & 5.8403  & 0.0996  & 0.2752  & 0.6409  & 23.0368  & 0.1742  & 0.0590  & 0.3337  & 0.3517  & 15.0281 & \textemdash & \textemdash  \\ 
    PCGen \cite{li2023pcgen}                 & 0.4636  & 0.8023  & 5.6583  & 0.2040  & 0.5391  & 0.4903  & 23.1675  & 0.1970  & 0.0763  & 0.5926  & 0.1351  & 14.1181 & \textemdash & \textemdash \\  
    \hline
    D-NeRF \cite{pumarola2021d}             & 0.1442  & 0.9128  & 4.0194  & 0.0508  & 0.3061  & 0.6634  & 26.2344  & 0.1369  & 0.0440  & 0.4309  & 0.3748  & 17.3554 & \textemdash & \textemdash  \\ 
    TiNeuVox-B \cite{fang2022fast}          & 0.1748  & 0.9059  & 4.1284  & 0.0502  & 0.3427  & 0.6514  & 26.0267  & 0.1363  & 0.0453  & 0.4365  & 0.3457  & 17.3535 & \textemdash & \textemdash  \\ 
    K-Planes \cite{fridovich2023k}           & 0.1302  & 0.9123  & 4.1322  & 0.0539  & 0.3457  & 0.6385  & 26.0236  & 0.1415  & 0.0498  & 0.4081  & 0.3008  & 17.0167 & \textemdash & \textemdash  \\ 
    \hline
    LiDAR-NeRF$^*$ \cite{tao2024lidar}           & 0.1438  & 0.9091  & 4.1753  & 0.0566  & 0.2797  & 0.6568  & 25.9878  & 0.1404  & 0.0443  & 0.3135  & 0.3831  & 17.1549 & 0.7500 & 0.3797  \\
    LiDAR4D$^*$ \cite{zheng2024lidar4d}          & 0.1089 & \textbf{0.9272} & 3.5256 & 0.0404 & 0.1051 & 0.7647 & 27.4767 & 0.1195 & 0.0327 & 0.1845 & 0.5304 & 18.5561 & 0.8080 & 0.5541 \\ 
    \midrule \midrule
    SN-LiDAR(Ours)                           & \textbf{0.0969} & 0.9269 & \textbf{2.9916} & \textbf{0.0359} & \textbf{0.0829} & \textbf{0.8601} & \textbf{28.8485} & \textbf{0.1073} & \textbf{0.0296} & \textbf{0.1593} & \textbf{0.6284} & \textbf{19.4351} & \textbf{0.8250} & \textbf{0.6159}\\ 
    \bottomrule
    \end{tabular}
    \begin{tablenotes}
        \footnotesize
        \item $^*$ means semantic metrics are obtained by pretrained segmentation model CENet\cite{cheng2022cenet}.
        \item Non-LiDAR methods are modified to LiDAR NVS pipeline.
    \end{tablenotes}
\end{threeparttable} 
}
\end{table*}
}
%%%%%%%%%%%%%%%%%%%%%%%%%%%%%%%%%%%%%%% Tables %%%%%%%%%%%%%%%%%%%%%%%%%%%%%%%%%%%%%%%%%%%

%%%%%%%%%%%%%%%%%%%%%%%%%%%%%%%%%%%%%%% Tables %%%%%%%%%%%%%%%%%%%%%%%%%%%%%%%%%%%%%%%%%%%
{\renewcommand{\arraystretch}{1.5}
\begin{table*}[!htp]
\centering
\caption{Quantitative comparison on SemanticKITTI dataset.}
\label{tab:quantity_semKITTI}
\resizebox{\textwidth}{!}{
\begin{threeparttable}
    \begin{tabular}{ccccccccccccccccccc}
    \toprule
    \multirow{2}{*}{Method}  & \multicolumn{2}{c}{Point Cloud} & \multicolumn{5}{c}{Depth} & \multicolumn{5}{c}{Intensity} & \multicolumn{3}{c}{Ray Drop} & \multicolumn{2}{c}{Semantic} \\ 
    \cmidrule(lr){2-3} \cmidrule(lr){4-8} \cmidrule(lr){9-13} \cmidrule(lr){14-16} \cmidrule(lr){17-18}
                           & CD$\downarrow$ & F-score$\uparrow$ & RMSE$\downarrow$ & MedAE$\downarrow$ & LPIPS$\downarrow$ & SSIM$\uparrow$ & PSNR$\uparrow$ & RMSE$\downarrow$ & MedAE$\downarrow$ & LPIPS$\downarrow$ & SSIM$\uparrow$ & PSNR$\uparrow$ & RMSE$\downarrow$ & Acc$\uparrow$ & F1-Score$\uparrow$ & PA$\uparrow$ & mIoU$\uparrow$ \\ 
    \midrule 
    LiDAR-NeRF$^*$ \cite{tao2024lidar}        & 0.1683  & 0.8833  & 4.7814  & 0.0795  & 0.2257  & 0.6418  & 24.9544  & 0.1464  & 0.0619  & 0.3751  & 0.2853  & 16.8306 & 0.3308 & 0.8604 & 0.9089 & 0.6465 & 0.3705 \\
    LiDAR4D$^*$ \cite{zheng2024lidar4d}       & \textbf{0.1175} & \textbf{0.9051} & 4.1070 & 0.0543 & 0.2125 & 0.7195 & 26.2559 & 0.1225 & 0.0421 & 0.2650 & 0.4370 & 18.3465 & 0.3038 & 0.8953 & 0.9345 & 0.8245 & 0.5323 \\ 
    \hline \hline
    SN-LiDAR(Ours)                        & 0.1236 & 0.8985 & \textbf{3.8619} & \textbf{0.0522} & \textbf{0.0900} & \textbf{0.8029} & \textbf{26.8046} & \textbf{0.1106} & \textbf{0.0371} & \textbf{0.1174} & \textbf{0.5469} & \textbf{19.2063} & \textbf{0.2357} & \textbf{0.9289} & \textbf{0.9544} & \textbf{0.9483} & \textbf{0.7904} \\ 
    \bottomrule
    \end{tabular}
    \begin{tablenotes}
        \footnotesize
        \item $^*$ means semantic metrics are obtained by pretrained segmentation model CENet\cite{cheng2022cenet}.
    \end{tablenotes}
\end{threeparttable} 
}
\end{table*}
}
%%%%%%%%%%%%%%%%%%%%%%%%%%%%%%%%%%%%%%% Tables %%%%%%%%%%%%%%%%%%%%%%%%%%%%%%%%%%%%%%%%%%%

\textbf{Reconstruction}. Tab. \ref{tab:quantity_KITTI360} and Tab. \ref{tab:quantity_semKITTI} present the quantitative comparisons on the KITTI-360 and SemanticKITTI datasets, respectively. Our  method demonstrates competitive results, outperforming previous methods across nearly all metrics. For geometric reconstruction, our depth achieves a 15\% and 5\% RMSE reduction on the two datasets compared to other methods, along with significant improvements in the quality of perception and structure (21\% and 57\%, 12\% and 11\%). Additionally, the accuracy of intensity and ray drop are also notably enhanced. These metrics highlight the positive impact of semantic understanding on geometric reconstruction, indicating that the joint optimization of semantic and geometric features enables the network to learn more accurate geometry.

\begin{figure*}[htp]
    \centering
    \includegraphics[width=\textwidth]{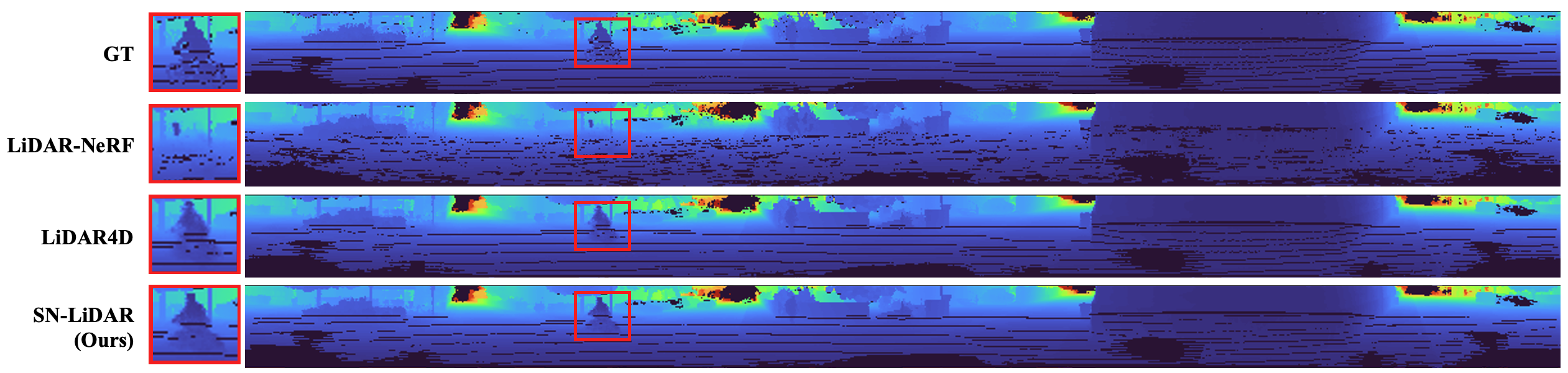}
    \caption{Qualitative comparison for LiDAR \textbf{depth} reconstruction and synthesis on SemanticKITTI. The red box shows the depth of the cyclist, with an enlarged view on the left.}
    \label{fig:depth_semkitti}
\end{figure*}

\begin{figure*}[htp]
    \centering
    \includegraphics[width=\textwidth]{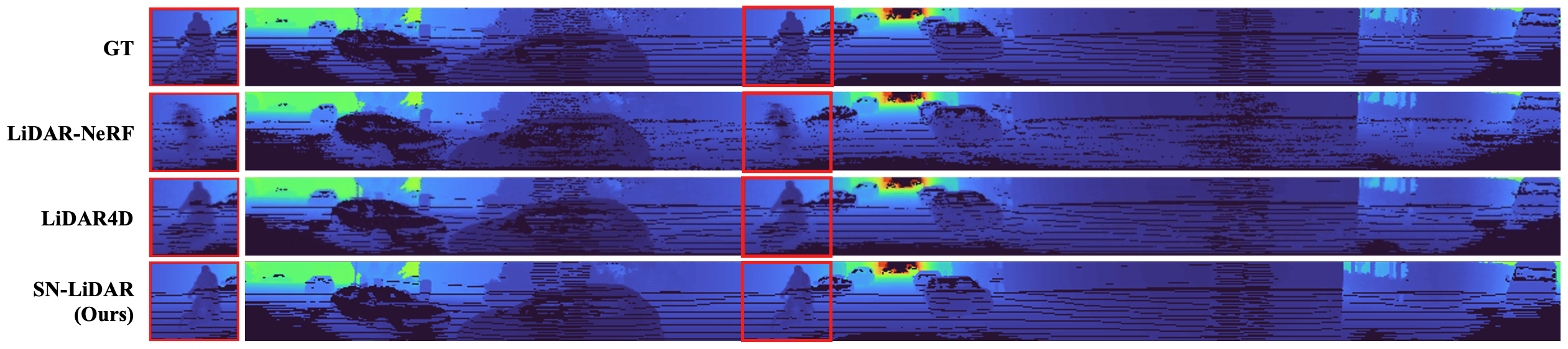}
    \caption{Qualitative comparison for LiDAR \textbf{depth} reconstruction and synthesis on KITTI-360. The red box shows the depth of the cyclist, with an enlarged view on the left.}
    \label{fig:depth_kitti360}
\end{figure*}

\begin{figure*}[htp]
    \centering
    \includegraphics[width=\textwidth]{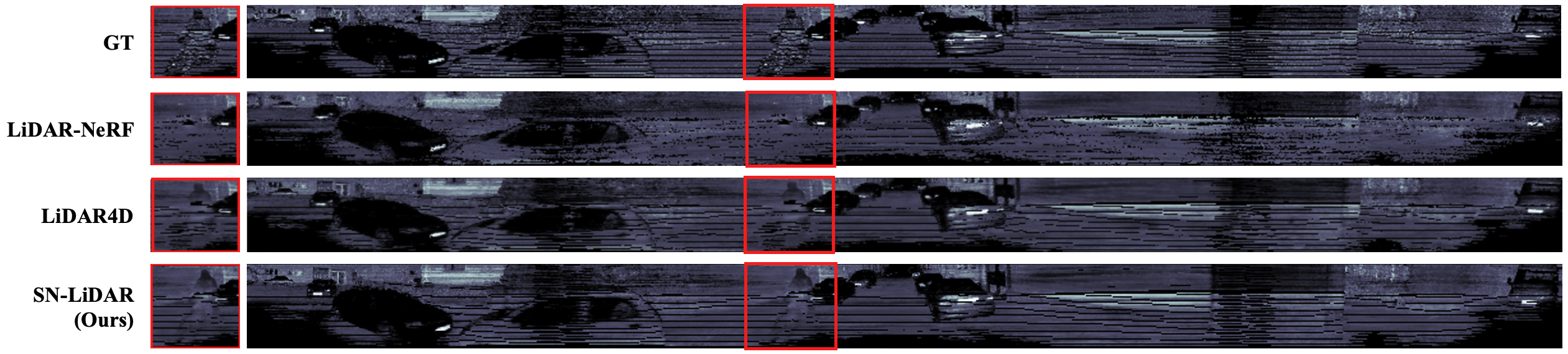}
    \caption{Qualitative comparison for LiDAR \textbf{intensity} reconstruction and synthesis on KITTI-360. The red box shows the intensity of the cyclist, with an enlarged view on the left.}
    \label{fig:intensity_kitti360}
\end{figure*}

However, our CD and F-score for point cloud reconstruction are slightly inferior to LiDAR4D. As shown in Fig. \ref{fig:pointcloud_semkitti_narrow}, our method tends to generate points in areas with gaps in the original point cloud to enhance depth smoothness and point cloud density, which results in some outliers that cause bad results of CD. For clearer visualization, we use a pre-trained semantic segmentation network to generate semantic predictions for point clouds synthesized by non-semantic methods. In Fig.~\ref{fig:pointcloud_semkitti_narrow}, we observe that our method excels at reconstructing small-sized dynamic objects. For example, in the red point cloud of pedestrians highlighted by the white box, both LiDAR-NeRF and LiDAR4D struggle to synthesize the legs, whereas SN-LiDAR successfully reconstructs the complete human form. As illustrated in Fig.~\ref{fig:depth_semkitti}, \ref{fig:depth_kitti360}, and \ref{fig:intensity_kitti360}, LiDAR-NeRF sometimes fails to reconstruct dynamic objects, whereas our method provides clearer boundary contours for moving pedestrians and bicycles compared to LiDAR4D. This demonstrates that, after the preliminary dynamic modeling by global planar-grid geometric features, our local CNN-based semantic encoder further enhances the fine-grained representation of small-sized objects.

%%%%%%%%%%%%%%%%%%%%%%%%%%%%%%%%%%%%%%% Tables %%%%%%%%%%%%%%%%%%%%%%%%%%%%%%%%%%%%%%%%%%%
{\renewcommand{\arraystretch}{1.4}
\begin{table}[!t]
\centering
\caption{Ablation Study.}
% GGR\mathcal{GGR}: Global Geometry Representation, SNF\mathcal{SNF}: Semantic Neural Fields, and LSR\mathcal{LSR}: Local Semantic Representation.
\label{tab:ablation}
\resizebox{0.5\textwidth}{!}{
\begin{tabular}{cccccccccccc}
\toprule
 \multirow{2}{*}{$\mathcal{GGR}$}  &  \multirow{2}{*}{$\mathcal{SNF}$} &  \multirow{2}{*}{$\mathcal{LSR}$} & \multicolumn{2}{c}{Point Cloud} & \multicolumn{5}{c}{Depth} & \multicolumn{2}{c}{Semantic} \\ 
\cmidrule(lr){4-5} \cmidrule(lr){6-10} \cmidrule(lr){11-12}
                      & & & CD$\downarrow$ & F-score$\uparrow$ & RMSE$\downarrow$ & MedAE$\downarrow$ & LPIPS$\downarrow$ & SSIM$\uparrow$ & PSNR$\uparrow$ & PA$\uparrow$ & mIoU$\uparrow$ \\ 
\midrule
\usym{2717} & \usym{2717} & \usym{2717}       & 0.4117 & 0.8350 & 6.4459  & 0.0908  & 0.2348 & 0.5862  & 21.9856 & 0.5870 & 0.3972  \\
$\checkmark$ & \usym{2717} & \usym{2717}      & 0.1491 & 0.8484 & 4.1779 & 0.0934 & 0.2019 & 0.6925 & 25.6592 & 0.8710 & 0.6798 \\ 
$\checkmark$ & $\checkmark$ & \usym{2717}     & 0.1459 & 0.8539 & 4.1124 & 0.0924 & 0.1986 & 0.7143 & 25.7980 & 0.9750 & 0.8844 \\ 
$\checkmark$ & $\checkmark$ & $\checkmark$    & \textbf{0.1308} & \textbf{0.8612} & \textbf{3.9173} & \textbf{0.0878} & \textbf{0.1180} & \textbf{0.7631} & \textbf{26.2046} & \textbf{0.9765} & \textbf{0.8923}\\ 
\bottomrule
\end{tabular}}
\end{table}
}
%%%%%%%%%%%%%%%%%%%%%%%%%%%%%%%%%%%%%%% Tables %%%%%%%%%%%%%%%%%%%%%%%%%%%%%%%%%%%%%%%%%%%

\begin{figure*}[htp]
    \centering
    \includegraphics[width=\textwidth]{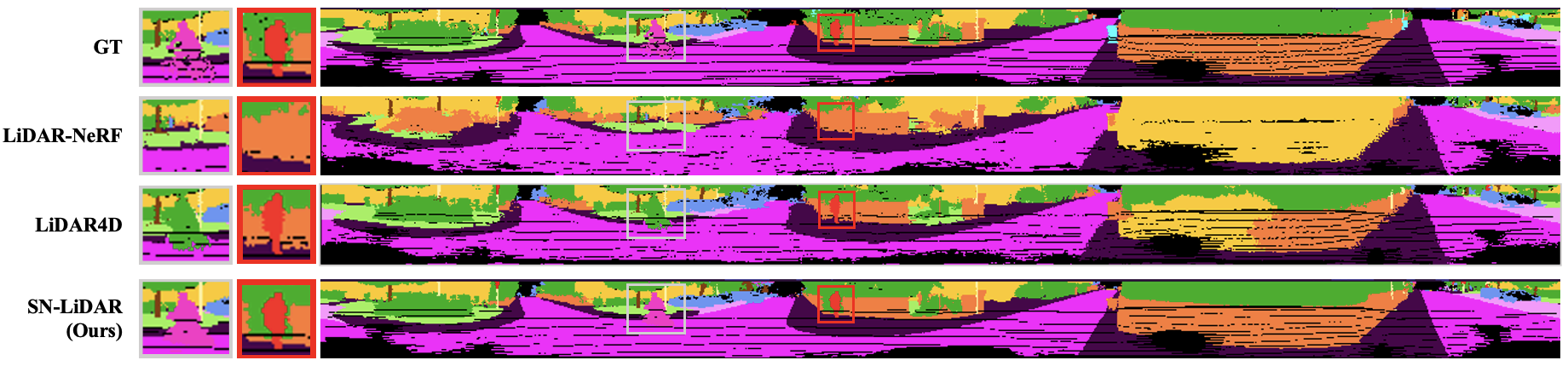}
    \caption{Qualitative comparison for LiDAR \textbf{semantic} reconstruction and synthesis on SemanticKITTI. The gray box displays the semantic label of the cyclist, and the red box shows the semantic label of the pedestrian, with an enlarged view on the left.}
    \label{fig:semantic_semkitti}
\end{figure*}

\textbf{Semantics}. To convincingly demonstrate the effectiveness of our semantic reconstruction, we employed CENet \cite{cheng2022cenet} pre-trained on SemanticKITTI to perform post-processing on the point clouds synthesized by other methods. Sharing the same structure as our local CNN-based encoder for a fair comparison, it was applied to perform semantic segmentation on point clouds synthesized by other methods, yielding the corresponding metrics. Both Tab.~\ref{tab:quantity_KITTI360} and Tab. \ref{tab:quantity_semKITTI}  show significant improvements in semantic reconstruction with our approach compared to the baseline methods. Fig. \ref{fig:semantic_semkitti} qualitatively compares the semantic reconstruction. The gray box highlights a cyclist in motion, and the red box features walking pedestrians. LiDAR-NeRF fails to reconstruct these dynamic objects; LiDAR4D manages to reconstruct the cyclist but produces incorrect semantics, and its reconstruction of dynamic objects is comparatively rough relative to our approach. These results further validate the mutual benefits between geometry and semantics of our method.

%%%%%%%%%%%%%%%%%%%%%%%%%%%%%%%%%%%%%%% Figures %%%%%%%%%%%%%%%%%%%%%%%%%%%%%%%%%%%%%%%%%%

%%%%%%%%%%%%%%%%%%%%%%%%%%%%%%%%%%%%%%% Figures %%%%%%%%%%%%%%%%%%%%%%%%%%%%%%%%%%%%%%%%%%

\subsection{Ablation Study}
We investigated the effectiveness of modules in SN-LiDAR. Tab.~\ref{tab:ablation} presents evaluations of point cloud reconstruction for \textbf{Global Geometry Representation} ($\mathcal{GGR}$), \textbf{Semantic Neural Fields} ({$\mathcal{SNF}$), and \textbf{Local Semantic Representation} ($\mathcal{LSR}$), noting that the data in this table were obtained using SemanticKITTI Sequence 05. The introduction of GGR resulted in a 63\% reduction in the CD of point cloud reconstruction, a 35\% decrease in RMSE of depth, and a 16\% increase in PSNR, demonstrating the effectiveness of this module in geometric perception. SNF significantly improved the PA and mIoU metrics for semantic reconstruction, proving the enhancement this module brings to semantic understanding. LSR showed improvements in point cloud CD, depth RMSE, and semantic mIoU, indicating that this module enhances local semantic details, thereby improving geometric and semantic reconstruction at the same time.

\section{Conclusion}
We propose SN-LiDAR, a semantic neural LiDAR fields that simultaneously performs accurate semantic segmentation, high-quality geometric reconstruction, and realistic LiDAR synthesis for novel space-time view. We combine global geometric and local semantic features within our local-to-global feature representation to enable mutual enhancement between geometry and semantics. Our experiments and evaluations on KITTI-360 and SemanticKITTI datasets demonstrate the superiority of our approach in semantic and geometric reconstruction. We hope that more future research will focus on novel LiDAR view synthesis with semantics.

\bibliographystyle{IEEEtran}
\bibliography{IEEEabrv,ref}

% Generated by IEEEtran.bst, version: 1.14 (2015/08/26)
\begin{thebibliography}{10}
\providecommand{\url}[1]{#1}
\csname url@samestyle\endcsname
\providecommand{\newblock}{\relax}
\providecommand{\bibinfo}[2]{#2}
\providecommand{\BIBentrySTDinterwordspacing}{\spaceskip=0pt\relax}
\providecommand{\BIBentryALTinterwordstretchfactor}{4}
\providecommand{\BIBentryALTinterwordspacing}{\spaceskip=\fontdimen2\font plus
\BIBentryALTinterwordstretchfactor\fontdimen3\font minus \fontdimen4\font\relax}
\providecommand{\BIBforeignlanguage}[2]{{%
\expandafter\ifx\csname l@#1\endcsname\relax
\typeout{** WARNING: IEEEtran.bst: No hyphenation pattern has been}%
\typeout{** loaded for the language `#1'. Using the pattern for}%
\typeout{** the default language instead.}%
\else
\language=\csname l@#1\endcsname
\fi
#2}}
\providecommand{\BIBdecl}{\relax}
\BIBdecl

\bibitem{dosovitskiy2017carla}
A.~Dosovitskiy, G.~Ros, F.~Codevilla, A.~Lopez, and V.~Koltun, ``Carla: An open urban driving simulator,'' in \emph{Conference on robot learning}.\hskip 1em plus 0.5em minus 0.4em\relax PMLR, 2017, pp. 1--16.

\bibitem{koenig2004design}
N.~Koenig and A.~Howard, ``Design and use paradigms for gazebo, an open-source multi-robot simulator,'' in \emph{2004 IEEE/RSJ international conference on intelligent robots and systems (IROS)(IEEE Cat. No. 04CH37566)}, vol.~3.\hskip 1em plus 0.5em minus 0.4em\relax Ieee, 2004, pp. 2149--2154.

\bibitem{manivasagam2020lidarsim}
S.~Manivasagam, S.~Wang, K.~Wong, W.~Zeng, M.~Sazanovich, S.~Tan, B.~Yang, W.-C. Ma, and R.~Urtasun, ``Lidarsim: Realistic lidar simulation by leveraging the real world,'' in \emph{Proceedings of the IEEE/CVF Conference on Computer Vision and Pattern Recognition}, 2020, pp. 11\,167--11\,176.

\bibitem{li2023pcgen}
C.~Li, Y.~Ren, and B.~Liu, ``Pcgen: Point cloud generator for lidar simulation,'' in \emph{2023 IEEE International Conference on Robotics and Automation (ICRA)}.\hskip 1em plus 0.5em minus 0.4em\relax IEEE, 2023, pp. 11\,676--11\,682.

\bibitem{pfister2000surfels}
H.~Pfister, M.~Zwicker, J.~Van~Baar, and M.~Gross, ``Surfels: Surface elements as rendering primitives,'' in \emph{Proceedings of the 27th annual conference on Computer graphics and interactive techniques}, 2000, pp. 335--342.

\bibitem{mildenhall2021nerf}
B.~Mildenhall, P.~P. Srinivasan, M.~Tancik, J.~T. Barron, R.~Ramamoorthi, and R.~Ng, ``Nerf: Representing scenes as neural radiance fields for view synthesis,'' \emph{Communications of the ACM}, vol.~65, no.~1, pp. 99--106, 2021.

\bibitem{prosgnerf}
T.~Deng, S.~Liu, X.~Wang, Y.~Liu, D.~Wang, and W.~Chen, ``Prosgnerf: Progressive dynamic neural scene graph with frequency modulated auto-encoder in urban scenes,'' \emph{arXiv preprint arXiv:2312.09076}, 2023.

\bibitem{plgslam}
T.~Deng, G.~Shen, T.~Qin, J.~Wang, W.~Zhao, J.~Wang, D.~Wang, and W.~Chen, ``Plgslam: Progressive neural scene represenation with local to global bundle adjustment,'' in \emph{Proceedings of the IEEE/CVF Conference on Computer Vision and Pattern Recognition}, 2024, pp. 19\,657--19\,666.

\bibitem{neslam}
T.~Deng, Y.~Wang, H.~Xie, H.~Wang, R.~Guo, J.~Wang, D.~Wang, and W.~Chen, ``Neslam: Neural implicit mapping and self-supervised feature tracking with depth completion and denoising,'' \emph{IEEE Transactions on Automation Science and Engineering}, 2025.

\bibitem{planning}
H.~Zhao, Z.~Ma, L.~Liu, Y.~Wang, Z.~Zhang, and H.~Liu, ``Optimized path planning for logistics robots using ant colony algorithm under multiple constraints,'' \emph{arXiv preprint arXiv:2504.05339}, 2025.

\bibitem{navigation}
Q.~Liu, H.~Xin, Z.~Liu, and H.~Wang, ``Integrating neural radiance fields end-to-end for cognitive visuomotor navigation,'' \emph{IEEE Transactions on Pattern Analysis and Machine Intelligence}, 2024.

\bibitem{huang2023neural}
S.~Huang, Z.~Gojcic, Z.~Wang, F.~Williams, Y.~Kasten, S.~Fidler, K.~Schindler, and O.~Litany, ``Neural lidar fields for novel view synthesis,'' in \emph{Proceedings of the IEEE/CVF International Conference on Computer Vision}, 2023, pp. 18\,236--18\,246.

\bibitem{tao2024lidar}
T.~Tao, L.~Gao, G.~Wang, Y.~Lao, P.~Chen, H.~Zhao, D.~Hao, X.~Liang, M.~Salzmann, and K.~Yu, ``Lidar-nerf: Novel lidar view synthesis via neural radiance fields,'' in \emph{Proceedings of the 32nd ACM International Conference on Multimedia}, 2024, pp. 390--398.

\bibitem{zheng2024lidar4d}
Z.~Zheng, F.~Lu, W.~Xue, G.~Chen, and C.~Jiang, ``Lidar4d: Dynamic neural fields for novel space-time view lidar synthesis,'' in \emph{Proceedings of the IEEE/CVF Conference on Computer Vision and Pattern Recognition}, 2024, pp. 5145--5154.

\bibitem{salt}
Y.~Wang, Y.~Chen, C.~Cao, T.~Deng, W.~Zhao, J.~Wang, and W.~Chen, ``Salt: A flexible semi-automatic labeling tool for general lidar point clouds with cross-scene adaptability and 4d consistency,'' \emph{arXiv preprint arXiv:2503.23980}, 2025.

\bibitem{sfpnet}
Y.~Wang, W.~Zhao, C.~Cao, T.~Deng, J.~Wang, and W.~Chen, ``Sfpnet: Sparse focal point network for semantic segmentation on general lidar point clouds,'' in \emph{European Conference on Computer Vision}.\hskip 1em plus 0.5em minus 0.4em\relax Springer, 2024, pp. 403--421.

\bibitem{zhi2021place}
S.~Zhi, T.~Laidlow, S.~Leutenegger, and A.~J. Davison, ``In-place scene labelling and understanding with implicit scene representation,'' in \emph{Proceedings of the IEEE/CVF International Conference on Computer Vision}, 2021, pp. 15\,838--15\,847.

\bibitem{vora2021nesf}
S.~Vora, N.~Radwan, K.~Greff, H.~Meyer, K.~Genova, M.~S.~M. Sajjadi, E.~Pot, A.~Tagliasacchi, and D.~Duckworth, ``Nesf: Neural semantic fields for generalizable semantic segmentation of 3d scenes,'' 2021.

\bibitem{cciccek20163d}
{\"O}.~{\c{C}}i{\c{c}}ek, A.~Abdulkadir, S.~S. Lienkamp, T.~Brox, and O.~Ronneberger, ``3d u-net: learning dense volumetric segmentation from sparse annotation,'' in \emph{Medical Image Computing and Computer-Assisted Intervention--MICCAI 2016: 19th International Conference, Athens, Greece, October 17-21, 2016, Proceedings, Part II 19}.\hskip 1em plus 0.5em minus 0.4em\relax Springer, 2016, pp. 424--432.

\bibitem{liu2023unsupervised}
Z.~Liu, F.~Milano, J.~Frey, R.~Siegwart, H.~Blum, and C.~Cadena, ``Unsupervised continual semantic adaptation through neural rendering,'' in \emph{Proceedings of the IEEE/CVF Conference on Computer Vision and Pattern Recognition}, 2023, pp. 3031--3040.

\bibitem{zhu2024sni}
S.~Zhu, G.~Wang, H.~Blum, J.~Liu, L.~Song, M.~Pollefeys, and H.~Wang, ``Sni-slam: Semantic neural implicit slam,'' in \emph{Proceedings of the IEEE/CVF Conference on Computer Vision and Pattern Recognition}, 2024, pp. 21\,167--21\,177.

\bibitem{sgsslam}
M.~Li, S.~Liu, H.~Zhou, G.~Zhu, N.~Cheng, T.~Deng, and H.~Wang, ``Sgs-slam: Semantic gaussian splatting for neural dense slam,'' in \emph{European Conference on Computer Vision}.\hskip 1em plus 0.5em minus 0.4em\relax Springer, 2024, pp. 163--179.

\bibitem{zhang2024nerf}
J.~Zhang, F.~Zhang, S.~Kuang, and L.~Zhang, ``Nerf-lidar: Generating realistic lidar point clouds with neural radiance fields,'' in \emph{Proceedings of the AAAI Conference on Artificial Intelligence}, vol.~38, no.~7, 2024, pp. 7178--7186.

\bibitem{muller2022instant}
T.~M{\"u}ller, A.~Evans, C.~Schied, and A.~Keller, ``Instant neural graphics primitives with a multiresolution hash encoding,'' \emph{ACM transactions on graphics (TOG)}, vol.~41, no.~4, pp. 1--15, 2022.

\bibitem{fridovich2023k}
S.~Fridovich-Keil, G.~Meanti, F.~R. Warburg, B.~Recht, and A.~Kanazawa, ``K-planes: Explicit radiance fields in space, time, and appearance,'' in \emph{Proceedings of the IEEE/CVF Conference on Computer Vision and Pattern Recognition}, 2023, pp. 12\,479--12\,488.

\bibitem{milioto2019rangenet++}
A.~Milioto, I.~Vizzo, J.~Behley, and C.~Stachniss, ``Rangenet++: Fast and accurate lidar semantic segmentation,'' in \emph{2019 IEEE/RSJ international conference on intelligent robots and systems (IROS)}.\hskip 1em plus 0.5em minus 0.4em\relax IEEE, 2019, pp. 4213--4220.

\bibitem{cheng2022cenet}
H.-X. Cheng, X.-F. Han, and G.-Q. Xiao, ``Cenet: Toward concise and efficient lidar semantic segmentation for autonomous driving,'' in \emph{2022 IEEE international conference on multimedia and expo (ICME)}.\hskip 1em plus 0.5em minus 0.4em\relax IEEE, 2022, pp. 01--06.

\bibitem{he2016deep}
K.~He, X.~Zhang, S.~Ren, and J.~Sun, ``Deep residual learning for image recognition,'' in \emph{Proceedings of the IEEE conference on computer vision and pattern recognition}, 2016, pp. 770--778.

\bibitem{howard2019searching}
A.~Howard, M.~Sandler, G.~Chu, L.-C. Chen, B.~Chen, M.~Tan, W.~Wang, Y.~Zhu, R.~Pang, V.~Vasudevan \emph{et~al.}, ``Searching for mobilenetv3,'' in \emph{Proceedings of the IEEE/CVF international conference on computer vision}, 2019, pp. 1314--1324.

\bibitem{behley2019iccv}
J.~Behley, M.~Garbade, A.~Milioto, J.~Quenzel, S.~Behnke, C.~Stachniss, and J.~Gall, ``{SemanticKITTI: A Dataset for Semantic Scene Understanding of LiDAR Sequences},'' in \emph{Proc. of the IEEE/CVF International Conf.~on Computer Vision (ICCV)}, 2019.

\bibitem{Liao2022PAMI}
Y.~Liao, J.~Xie, and A.~Geiger, ``{KITTI}-360: A novel dataset and benchmarks for urban scene understanding in 2d and 3d,'' \emph{Pattern Analysis and Machine Intelligence (PAMI)}, 2022.

\bibitem{fan2017point}
H.~Fan, H.~Su, and L.~J. Guibas, ``A point set generation network for 3d object reconstruction from a single image,'' in \emph{Proceedings of the IEEE conference on computer vision and pattern recognition}, 2017, pp. 605--613.

\bibitem{wang2004image}
Z.~Wang, A.~C. Bovik, H.~R. Sheikh, and E.~P. Simoncelli, ``Image quality assessment: from error visibility to structural similarity,'' \emph{IEEE transactions on image processing}, vol.~13, no.~4, pp. 600--612, 2004.

\bibitem{zhang2018unreasonable}
R.~Zhang, P.~Isola, A.~A. Efros, E.~Shechtman, and O.~Wang, ``The unreasonable effectiveness of deep features as a perceptual metric,'' in \emph{Proceedings of the IEEE conference on computer vision and pattern recognition}, 2018, pp. 586--595.

\bibitem{long2015fully}
J.~Long, E.~Shelhamer, and T.~Darrell, ``Fully convolutional networks for semantic segmentation,'' in \emph{Proceedings of the IEEE conference on computer vision and pattern recognition}, 2015, pp. 3431--3440.

\bibitem{huang2023nksr}
J.~Huang, Z.~Gojcic, M.~Atzmon, O.~Litany, S.~Fidler, and F.~Williams, ``Neural kernel surface reconstruction,'' in \emph{Proceedings of the IEEE/CVF Conference on Computer Vision and Pattern Recognition}, 2023, pp. 4369--4379.

\bibitem{pumarola2021d}
A.~Pumarola, E.~Corona, G.~Pons-Moll, and F.~Moreno-Noguer, ``D-nerf: Neural radiance fields for dynamic scenes,'' in \emph{Proceedings of the IEEE/CVF conference on computer vision and pattern recognition}, 2021, pp. 10\,318--10\,327.

\bibitem{fang2022fast}
J.~Fang, T.~Yi, X.~Wang, L.~Xie, X.~Zhang, W.~Liu, M.~Nie{\ss}ner, and Q.~Tian, ``Fast dynamic radiance fields with time-aware neural voxels,'' in \emph{SIGGRAPH Asia 2022 Conference Papers}, 2022, pp. 1--9.

\end{thebibliography}

\end{document}